\pdfoutput=1

\documentclass[11pt]{article}

\usepackage[final]{acl}

\usepackage{times}
\usepackage{latexsym}
\usepackage{comment}
\usepackage{booktabs}

\usepackage[T1]{fontenc}

\usepackage[utf8]{inputenc}

\usepackage{microtype}

\usepackage{inconsolata}

\usepackage{graphicx}
\usepackage{tabularx}
\usepackage{listings}
\lstdefinelanguage{json}{
    basicstyle=\small\ttfamily, 
    stepnumber=1,
    numbersep=8pt,
    showstringspaces=false,
    breaklines=true,
    frame=lines
}
\usepackage[most]{tcolorbox}
\newtcolorbox{promptbox}{
    enhanced,
    colback=gray!5,
    colframe=gray!50,
    boxrule=0.5pt,
    fontupper=\small,
    arc=2pt,
    outer arc=2pt,
    boxsep=5pt,
    left=8pt,
    right=8pt,
    top=6pt,
    bottom=6pt
}

\newcommand\blfootnote[1]{%
  \begingroup
  \renewcommand\thefootnote{}%
  \footnote{#1}%
  \addtocounter{footnote}{-1}%
  \endgroup
}

\title{RouteNator:\\
A Router-Based Multi-Modal Architecture for Generating Synthetic Training Data for Function Calling LLMs}

\author{
Vibha Belavadi$^{*}$, Tushar Vatsa$^{*}$, Dewang Sultania$^{*}$, Suhas Suresha$^{*}$,  Ishita Verma$^{*}$,\\ {\bf Cheng Chen}, {\bf Tracy Holloway King}, {\bf Michael Friedrich} \\
Adobe Inc. \\
 345 Park Avenue\\
    San Jose, CA, 95110\\
}

\begin{document}
\maketitle

\blfootnote{$^*$ These authors contributed equally to this work.}

\begin{abstract}
This paper addresses fine-tuning Large Language Models (LLMs) for function calling tasks when real user interaction data is unavailable. In digital content creation tools, where users express their needs through natural language queries that must be mapped to API calls, the lack of real-world task-specific data and privacy constraints for training on it necessitate synthetic data generation. Existing approaches to synthetic data generation fall short in diversity and complexity, failing to replicate real-world data distributions and leading to suboptimal performance after LLM fine-tuning. We present a novel router-based architecture that leverages domain resources like content metadata and structured knowledge graphs, along with text-to-text and vision-to-text language models to generate high-quality synthetic training data. Our architecture's flexible routing mechanism enables synthetic data generation that matches observed real-world distributions, addressing a fundamental limitation of traditional approaches. Evaluation on a comprehensive set of real user queries demonstrates significant improvements in both function classification accuracy and API parameter selection. Models fine-tuned with our synthetic data consistently outperform traditional approaches, establishing new benchmarks for function calling tasks. 
\end{abstract}

\section{Introduction}

Digital content creation platforms increasingly rely on natural language interfaces to make complex design tools accessible to non-technical users. A critical challenge lies in accurately translating user queries into appropriate function calls \cite{c35} for instance, when a user requests ``Find me an image of an elephant with the background being Taj Mahal'', the system must orchestrate multiple API calls for searching, background removal, and compositing.

In this paper, we specifically address the challenge of training models to classify user intent into two distinct categories: queries that can be fulfilled through search API operations versus those requiring generation through Generative-AI-powered APIs. Given a user query, our model determines: (1) whether to route the request to ``Search'' or ``Generate'' operations based on user intent, (2) the appropriate Content Type parameter selection (e.g.\ Photo, Template, Background, Video), and (3) prompt optimization specific to each API type—simplifying ``Search'' queries while preserving detailed specifications for ``Generate'' API requests. The examples below in Listing~\ref{lst:function_calling} demonstrate how our model processes and classifies different types of user queries:

\begin{lstlisting}[language=json,firstnumber=1,caption=Example function calls for  user queries]
example1 = {
'input': 'Find me an image of an elephant',
'output': {
   'function': 'Search',
   'content_type': 'Photo',
   'extracted_prompt': 'elephant' 
   } 
},
example2 = {
'input': 'Create a birthday invitation for my nephew whose birthday is on January 21',
'output': {
    'function': 'Generate',
    'content_type': 'Template',
    'extracted_prompt': 'invitation for nephew's birthday on January 21' 
   } 
}
\end{lstlisting}
\label{lst:function_calling}

\begin{figure*}[t]
    \centering\small
    \includegraphics[width=0.4\textwidth]{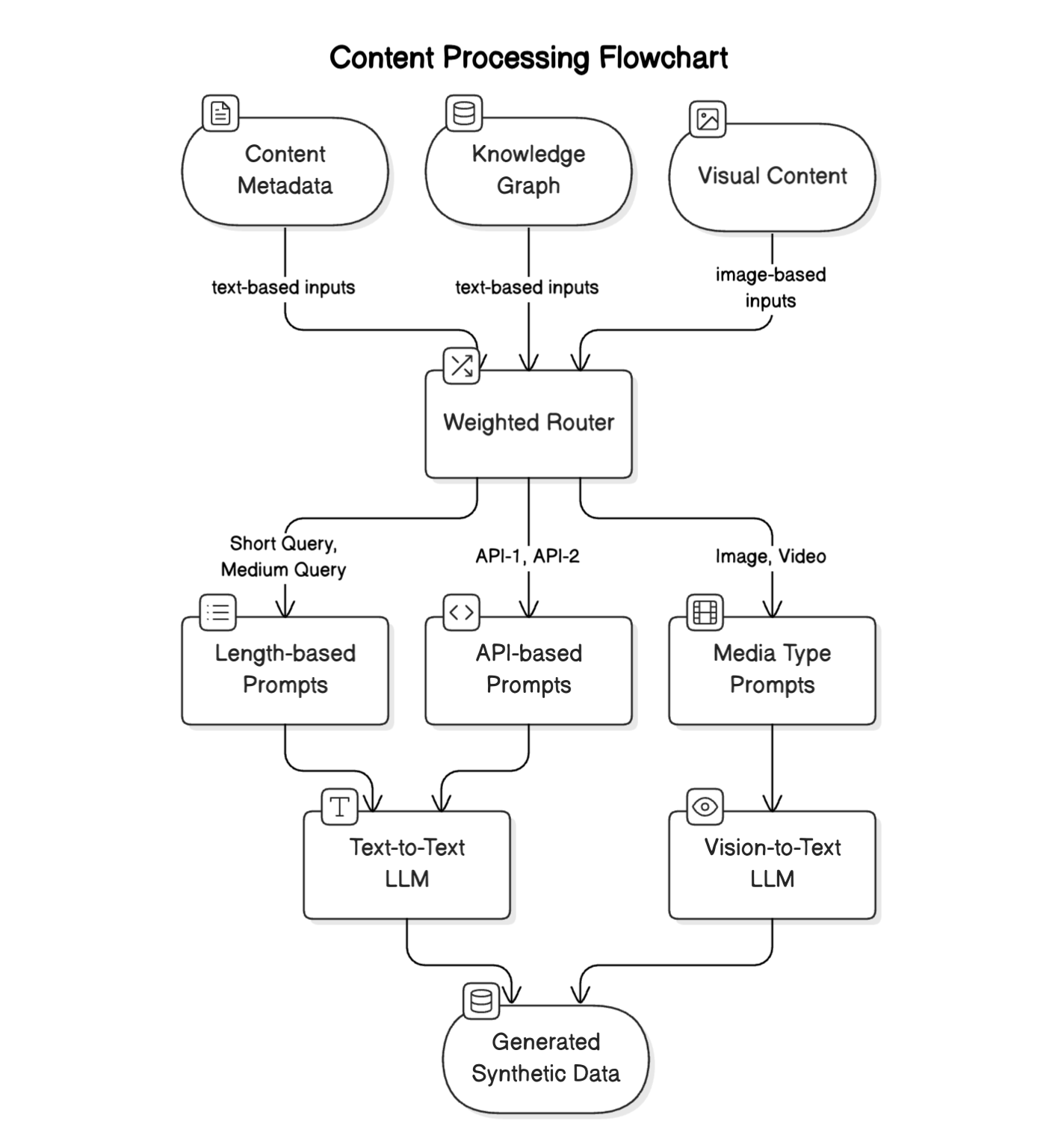}
    \caption{Data generation architecture overview integrating metadata, knowledge graph, and visual content. A ``Weighted Router'' directs text and image inputs to different prompt categories: length-based, API-based, and media type. They are processed by Text-to-Text and Vision-to-Text LLMs to generate synthetic data for downstream tasks.}
    \label{fig:architecture}
\end{figure*}

While existing function calling models \cite{c22} show promise, their performance on specialized domains remains suboptimal and privacy restrictions on production data create training challenges. To address these limitations, we present two key contributions:

1. A methodology for incorporating structured domain knowledge into synthetic data generation that leverages: (a) Techniques for extracting generalizable patterns from content metadata; (b) Methods for utilizing domain-specific knowledge graphs to generate contextually relevant queries.

2. A novel router-based architecture for synthetic data generation featuring: (a) Multiple specialized LLM prompt templates as distinct routes; 
(b) A weighted routing mechanism using population-level statistics;

(c) Integration of multi-modal language models to increase data diversity.

Our approach (Figure~\ref{fig:architecture}) improves downstream model performance while producing balanced training data across content types, with well-distributed keywords and diverse sentence structures that better align with real-world user interactions.

\section{Related Work}

The challenge of generating high-quality synthetic training data for language models has been explored through various approaches. Prior work in generating high-quality synthetic training data for language models spans three key categories:
Instruction-tuning approaches have shown significant promise, starting with Self-Instruct's \cite{c11} 175-seed task framework. The field expanded through WizardLM's EvolInstruct \cite{c13}, Unnatural Instructions \cite{c12}, FLAN \cite{c14}, FLAN-T5 \cite{c15}, Alpaca \cite{c16}, Prompt-Breeder \cite{c17}, and Template-based Generation \cite{c18}.
Multimodal synthetic data generation advanced through Visual Instruction Tuning \cite{c19}, MiniGPT-4 \cite{c20}, and InternVL \cite{c21}, incorporating visual and textual information for enhanced data generation.
Function calling approaches, exemplified by Gorilla \cite{c22} building on Self-Instruct \cite{c11}, addressed API parameter matching challenges, though lacking domain-specific knowledge integration. 


Our work differs from previous approaches in several key aspects. (1) We focus specifically on generating synthetic data for function calling while maintaining real-world query distributions. Unlike general instruction tuning approaches, we  target the unique challenges of function calling data generation which include  precise parameter matching requirements, maintaining real-world API usage distributions and handling complex nested function calls. (2) We introduce a novel router-based architecture that combines multiple generation strategies. We extend existing router-based approaches by adding weighted probabilistic sampling and by using population-level statistics to guide routing decisions. We also combine text-to-text and vision-to-text generation paths. (3) We incorporate domain-specific knowledge while respecting privacy constraints by not directly referring to the real-world datasets. (4) We leverage multiple modalities (text and images) to increase the diversity and quality of generated data, particularly for visual content-related APIs. We introduce novel evaluation metrics measuring the qualitative alignment of the synthetic data with real-world data covering content-type alignment, diversity in data types generated, word length variability, and positional variance of key terms within sentences. 

\section{Methodology and Experiments}

This section details how our data generation approach evolved.

\subsection{Template-based Heuristic Generation}

\begin{figure}[ht!]
    \centering\small
    \includegraphics[width=0.45\textwidth]{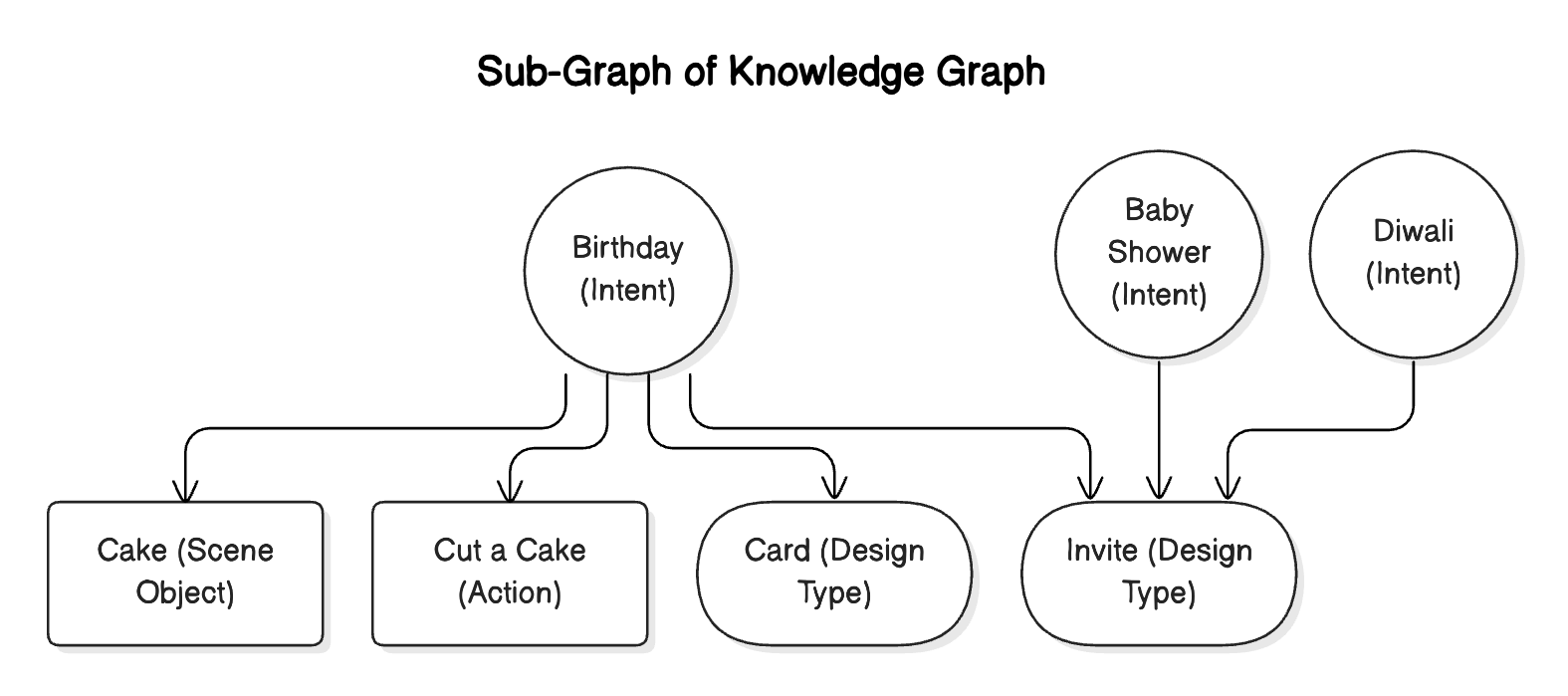}
    \caption{Knowledge Graph of concepts linked by edges}
    \label{fig:knowledge_graph}
\end{figure}

Our initial approach employed rule-based templates that combined content metadata with domain-specific Knowledge Graph (KG) relationships \cite{c23} between different aspects of digital content creation. Consider a snapshot of the Knowledge Graph sub-graph (Figure~\ref{fig:knowledge_graph}). This sub-graph consists of interconnected nodes representing User Intents ("Birthday", "Diwali", "Baby shower"), Design Types ("Card", "Invite"), Scene Objects ("Cake"), and associated Actions ("Cut a cake"). Each edge between the nodes represents a relationship between them. These connections were created using the historical and semantic relationships seen between different entities like User Intents, Design Types, Scene Objects and Actions. They enable the generation of semantically coherent queries by following established relationships between concepts. We create synthetic ``Search'' API data by generating random prompts combining the related entities (e.g.\ Intent and Design Type) with search synonyms e.g.\ "find me", "search for", "look for", "search", "show me". 

\begin{lstlisting}[language=json, caption=Examples of Image and Template metadata]
image_asset_metadata = {
...
'title': 'Tropical frangipani flowers floating',
'keywords': ['flower', 'frangipani', 'paradise', 'turquoise', 'tranquil', 'tropical', 'summer'],
'gentech': False
...
},

template_asset_metadata = {
...
'topics': ['galactic','space','server banner']
'title': 'Galaxy Minecraft Server Banner',
...
}
\end{lstlisting}
\label{lst:content_metadata}

In addition to using the Knowledge Graph, we also use content metadata of templates and images to heuristically create data. Each image or template asset contains metadata capturing its characteristics, such as the title of the asset, keywords or tags associated with it, whether it was generated by AI, locale, aspect ratio, click through rate, etc. This metadata provides a foundation of contextually relevant information that reflects real-world content organization and classification. Listing~2 
captures some of the image and template metadata tags used. For ``Generate'' API queries, we constructed templates that combined action verbs (e.g.\ "generate", "create" or "make") with content design types and titles from our metadata. For example, a template might expand to ``please generate a template for" followed by the title from our content metadata. For creating ``Generate'' API queries for Image assets, we only use the asset if the label `gentech' is set to True. Similarly, for ``Search'' API queries, we used search-related verbs (e.g.\ "find", "search for") with appropriate content descriptors. 
This approach allowed rapid generation of synthetic data with  proportions matching real-world statistics, but suffered from significant limitations: The generated queries lacked diversity, often with unnatural language patterns.

\subsection{Single-Prompt LLM Based Generation}

To address these limitations, we experimented with  a Llama-3.1-70B-Instruct model \cite{c25} with a comprehensive set of prompts containing API specifications and few-shot examples.  Different variants of the system prompt focused on different aspects of the content metadata e.g.\ intents, assets, actions. Examples of the Llama model prompts used used for synthesizing ``Search'' and ``Generate'' API queries are listed in appendix~\ref{app:appendix_llama_model_prompts}. To mimic the characteristics of real-world data (e.g.\ query length for ``Search'' vs ``Generate'' queries), the system prompts used for ``Search'' queries specify the query to be short and crisp. Conversely the ``Generate'' query prompts used layout creativity and engagement as a driving factor for data generation. This approach generated more natural language queries but presented challenges in controlling output distributions and maintaining variety across generated samples. Furthermore, it was difficult to ensure appropriate coverage across different content types and query patterns. 

In addition to Llama 70B model, a key innovation in our approach is  the integration of multi-modal capabilities for synthetic data generation on template data through the InternVL vision-to-text model \cite{c21}. We prompted the 40B InternVL model to generate a few queries that would result in the creation of the input template placing emphasis on the important elements unique to the template. The prompt for InternVL model is shared in Appendix \ref{app:appendix_internvl_model_prompts}. This addition of multi-modality based data generation component enables the generation of queries based on actual domain-specific corpus images and visual representations of non-image content. This provides an additional route for query generation that captures visual aspects that are not present in the metadata, leading to more natural descriptions and increased output diversity. 

\subsection{Router-based Multi-Modal Architecture}

Our final approach introduced a novel router-based architecture that  addresses the limitations of the previous two methods. The architecture consists of multiple specialized prompt templates, each designed to generate specific types of queries based on length, API type, and content requirements. These prompt templates incorporate variables from content metadata and domain-specific KG \cite{c23} relationships, ensuring semantic relevance while maintaining natural language patterns. The architecture also employs dataset generation from the approaches discussed above: heuristic-based, Llama text-to-text model and InternVL's vision-to-text model. 

The core of our architecture is a weighted router that directs query generation requests to appropriate prompt templates based on population-level statistics. This routing mechanism implements weighted sampling to maintain realistic query patterns while ensuring coverage across different query types and content categories. Table~\ref{tab:data_gen_nums} gives the distribution of the synthetic dataset generated across the heuristic-, single-prompt- and router-based approaches. 

\begin{table}[t]
\centering\small
\begin{tabular}{lcc}
\toprule
\textbf{Synthetic Dataset variant} & \textbf{Search} & \textbf{Generate} \\ \midrule
Heuristic Based & 103,189 & 102,922\\ 
Single-Prompt LLM Based & 100,207 & 100,433\\
Router-Based & 105,100 & 110,000\\ 
\bottomrule
\end{tabular}
\caption{Number of synthetic training examples generated for Search and Generate functions for each data generation approach}
\label{tab:data_gen_nums}
\end{table}

\subsection{Implementation Details}

The router selection algorithm  determines target distributions based on population statistics and selects prompt templates based on required query characteristics and content type requirements. For each synthetic data point, the router either selects a text-based route, populating templates with metadata and KG  elements, or a vision-based route, processing content images through InternVL to generate contextually relevant queries. 

The system includes  validation checks for query realism, label accuracy, and distribution alignment. It filters out duplicate queries, unrealistic language patterns, and queries that violate length constraints. This ensures that the  synthetic data is high quality and accurately reflects real-world usage patterns. 

The query generation process is continuously monitored and adjusted to maintain desired distributions across query lengths, API usage patterns, and content type frequencies. This adaptive approach ensures that the generated dataset remains balanced and representative of current user behavior patterns, while the multi-modal integration provides  diversity and realism in the generated queries. 

\begin{figure}[ht!]
    \centering\small
    \includegraphics[width=0.48\textwidth]{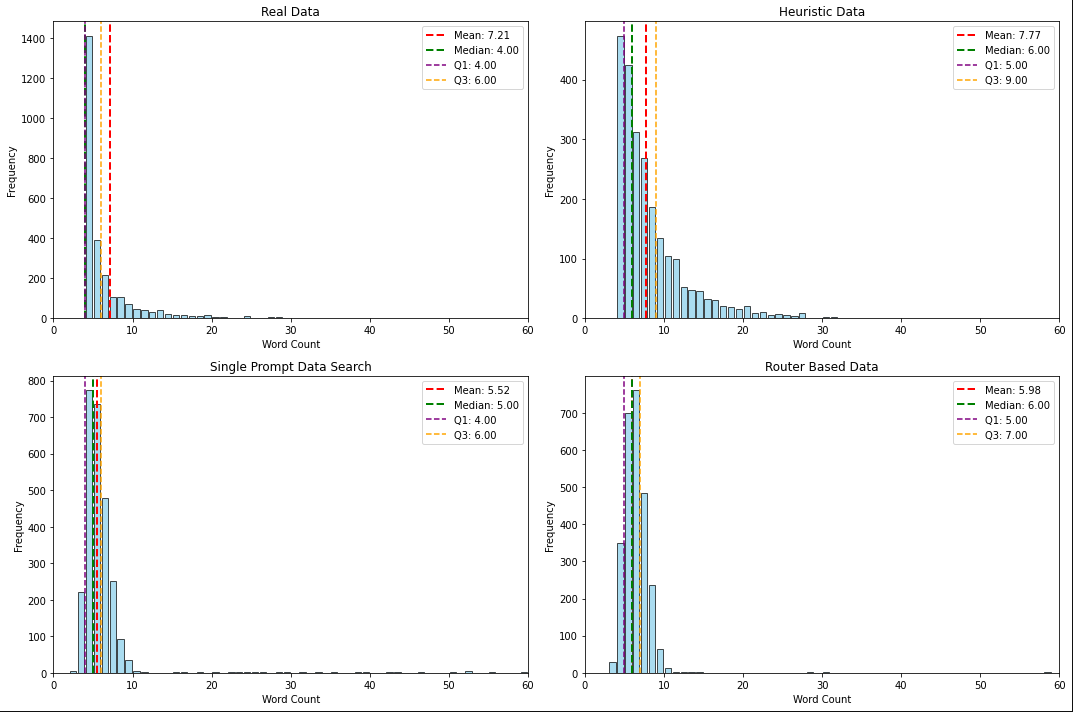}
    \caption{Comparison of word count distribution (Mean, Median and Interquartile Range) across the real and synthetically generated datasets (Heuristic, Single Prompt and Router)}
    \label{fig:word_count_distribution_diversity}
\end{figure}

\subsection{Fine-tuning}

To efficiently adapt the models while managing computational resources, we employ Quantized Low-Rank Adaptation (QLoRA)~\cite{c26} across our experiments. For the Gorilla OpenFunctions v2 model \cite{c22}, we utilize 4 NVIDIA A100 GPUs operating in parallel, with the base model parameters quantized to 4-bit precision while maintaining model quality through low-rank adapters. The training configuration utilizes cosine annealing for learning rate optimization \cite{c27} with ADAM\_W \cite{c28} as the optimizer, and we set the LoRA rank and alpha parameters to 16 and 32 respectively to balance adaptation capability with training stability.

Additionally, we fine-tuned several small language models (SLMs) using the same QLoRA technique Phi-3.5-mini-instruct \cite{c32}, Llama-3.2-1B-Instruct \cite{c33}, Llama-3.2-3B-Instruct \cite{c33}, Qwen2.5-1.5B-Instruct \cite{c31}, Qwen2.5-0.5B-Instruct \cite{c31} and Gemma-2-2b-it \cite{c30}. For these models, the training infrastructure consisted of 4 NVIDIA A10 GPUs operating in parallel. We maintained consistent quantization and adaptation strategies across all models to ensure fair comparison. The  hyperparameters and prompt structure used for training, training and evaluation loss, system memory usage and GPU utilization are  in Appendix~\ref{app:appendix_model_finetuning}.

\section{Results and Analysis}
We analyze the results by first looking at the data diversity of the router based synthetic data (word count distribution, content type diversity, positional diversity of keywords and query length distribution) and comparing it with other synthetically generated datasets. We then focus on the performance metrics of different variants of Gorilla model across different synthetic datasets. We also show the performance improvement of Small Language Models (SLMs) fine-tuned on our router based synthetic dataset compared to their base model.


\subsection{Word Count Distribution}

\begin{table*}[!t]
\centering\small
\makebox[\textwidth]{
\begin{tabular}{p{6cm}p{1.5cm}p{2cm}p{4.5cm}}
\toprule
\textbf{User Query} & \textbf{Function} & \textbf{Content Type} & \textbf{Optimized Subprompt} \\
\midrule
Find me a birthday template with balloons and confetti & Search & Template & birthday balloons confetti \\
\midrule
Create an elegant wedding invitation with gold floral borders for a December ceremony & Generate & Template & elegant wedding invitation gold floral borders December ceremony \\
\midrule
Show me tropical beach backgrounds & Search & Background & tropical beach background \\
\midrule
Generate a podcast cover with neon colors and retro style & Generate & Design Asset & podcast cover neon colors retro style \\
\midrule
Find business presentation templates with data charts & Search & Template & business presentation data charts \\
\bottomrule
\end{tabular}}
\caption{Representative Examples from Golden Dataset}
\label{tab:example-queries}
\end{table*}

Figure~\ref{fig:word_count_distribution_diversity}  compares the word count distribution across real-world and synthetically generated datasets, specifically analyzing the mean, median, and Interquartile Range (IQR). For this comparison, we sampled 2,500 search queries from each distribution.

The real-world dataset has a mean word length of 7 words and a median of 4 words. The distribution is right-skewed with a short IQR, suggesting that real-world queries are generally concise, typically ranging between 1 to 10 words.

In contrast, the synthetic dataset generated using Heuristic Data (KG and metadata-based) exhibits a higher median of 6 words, indicating that the generated queries tend to be more verbose. Although the Single Prompt-based data has a similar IQR, its narrower distribution suggests that the synthetic queries are, on average, shorter than real-world queries.

Finally, we observe that the Router-based synthetic data generation approach maintains a similar IQR to the real-world data, while  achieving a balanced distribution between diverse and realistic queries. This means that the generated queries are neither excessively long nor too short, aligning with real-world user behavior—where users are likely to search with either "Search" (short user query) or "Generate" (long user query).

\subsection{Content Type Diversity}

\begin{figure}[h!]
    \centering\small
    \includegraphics[width=0.4\textwidth]{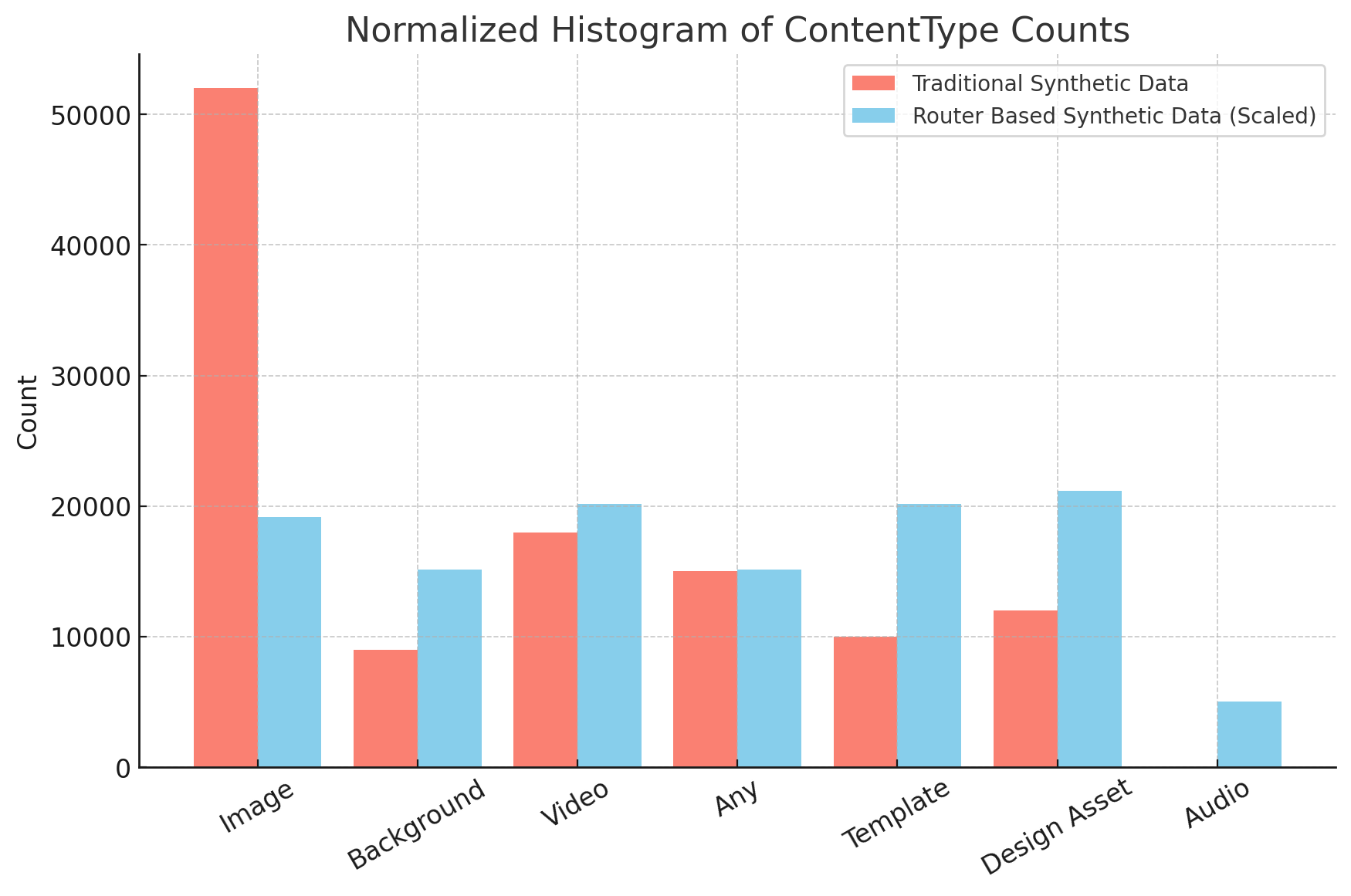}
    \caption{Comparison of Content Type distribution}
    \label{fig:content-bias}
\end{figure}

Traditional synthetic data generation techniques often struggle to replicate a real-world data diversity, resulting in imbalanced datasets where certain content types are overrepresented. In contrast, our  architecture enables a balanced and diverse distribution across content types.
 Figure~\ref{fig:content-bias} shows that our approach achieves a relatively even distribution across multiple content types (e.g.\ `Templates', `Images', `Videos', `Backgrounds'), allowing the model to learn  from a variety of content requests without over-fitting to any single category. This balanced distribution ensures that the model is exposed to a realistic sampling of potential queries, improving its generalization ability for content-specific API calls.
In contrast, traditional synthetic data generation methods (Figure~\ref{fig:content-bias}) tend to be heavily skewed, with content types like `Image' dominating the dataset, while others such as `Audio' and `Template' are underrepresented. This can limit a model’s capability to handle less frequent but important content types, resulting in suboptimal performance in real-world applications.

\begin{table*}[t]
\centering\small
\begin{tabular}{lccc}
\toprule
\textbf{Model and Dataset variant} & \textbf{Function Call} & \textbf{ContentType} & \textbf{Subprompt} \\
\textbf{} & \textbf{F1-Score} & \textbf{ Accuracy (CTA)} & \textbf{Similarity (SS)} \\ \midrule
Pre Fine-tuned Base Gorilla & 0.646 & 0.239 & 0.824 \\ 
Fine-tuned Gorilla: Single Prompt dataset & 0.788 & 0.574 & 0.898 \\
Fine-tuned Gorilla: Heuristic dataset & 0.801 & 0.676 & \textbf{0.919} \\ 
Fine-tuned Gorilla: Synthetic dataset + Router & 0.844 & 0.65 & 0.867 \\ 
Fine-tuned Gorilla: Synthetic + Heuristic dataset + Router & 0.875 & 0.737 & 0.915 \\ 
Prompt Tuned Gorilla: Synthetic + Heuristic dataset + Router& \textbf{0.881} & \textbf{0.756} & 0.918 \\ 
\bottomrule
\end{tabular}
\caption{Performance summary of the fine-tuned Gorilla model trained on different datasets. The ContentType Accuracy and Subprompt Similarity are referenced as CTA and SS respectively}
\label{tab:model_performance}
\end{table*}

\subsection{Positional Diversity of Keywords}
One of the key improvements in our synthetic data generation approach is the reduction of keyword position bias, specifically for `Content Type' keywords within user queries. Traditional synthetic datasets often position these keywords (e.g.\ `Image', `Video', `Template', `Audio') consistently at the beginning or end of queries. This lack of positional diversity leads to models that are prone to over-fitting, as they learn to expect keywords in fixed positions, which limits their generalization capabilities in real-world scenarios.

\begin{figure}[h!]
    \centering\small
    \includegraphics[width=0.4\textwidth]{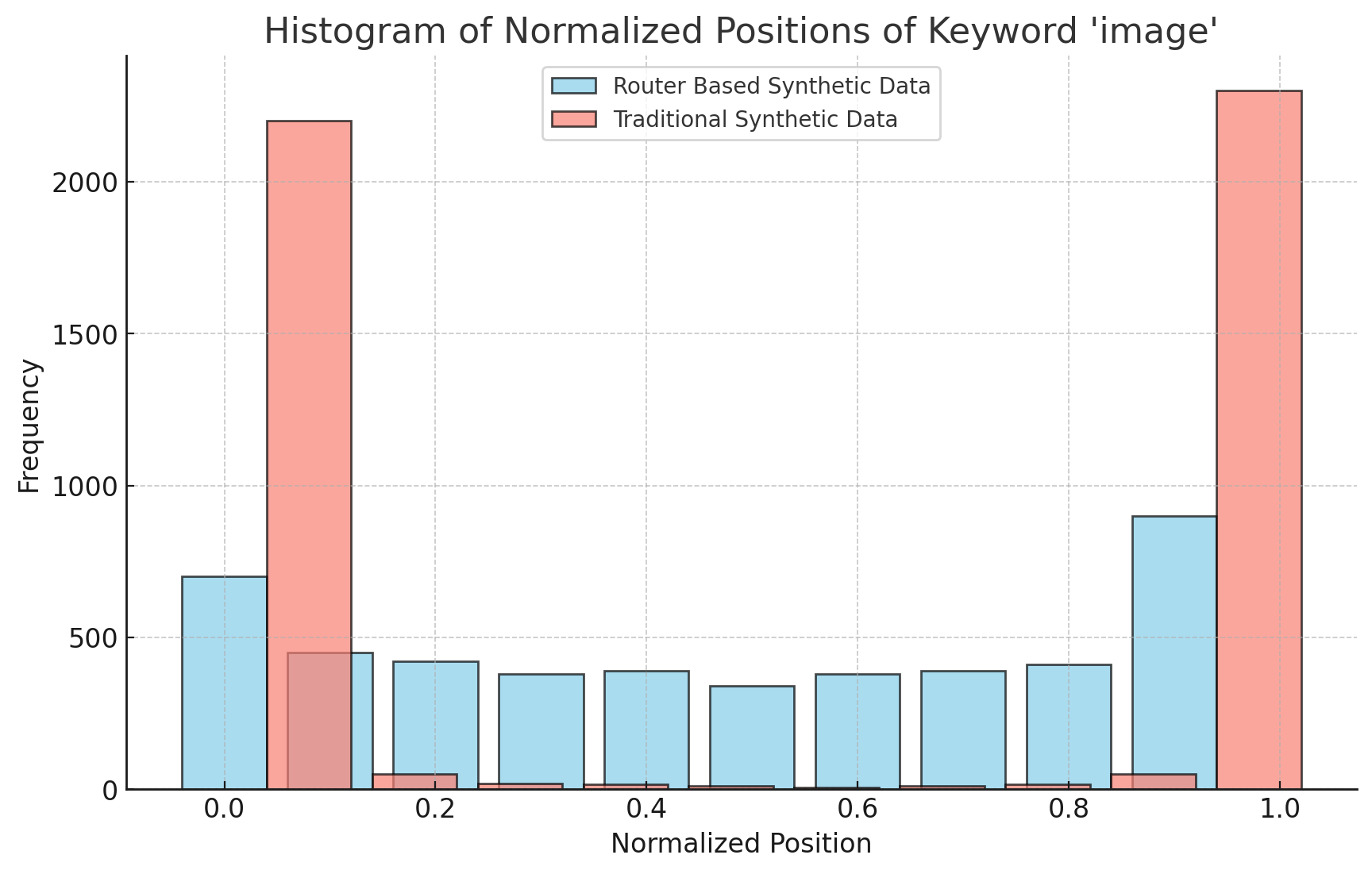}
    \caption{Comparison of normalized keyword positions}
    \label{fig:position_bias}
\end{figure}

Our router-based synthetic data generation framework creates a more even distribution of content type keywords (Figure~\ref{fig:position_bias}) across  positions in the query, which exposes the model to a wider range of query structures, helping it generalize  and reducing over-fitting. 

\subsection{Query Length Diversity}
\begin{figure}[h!]
    \centering
    \includegraphics[width=0.45\textwidth]{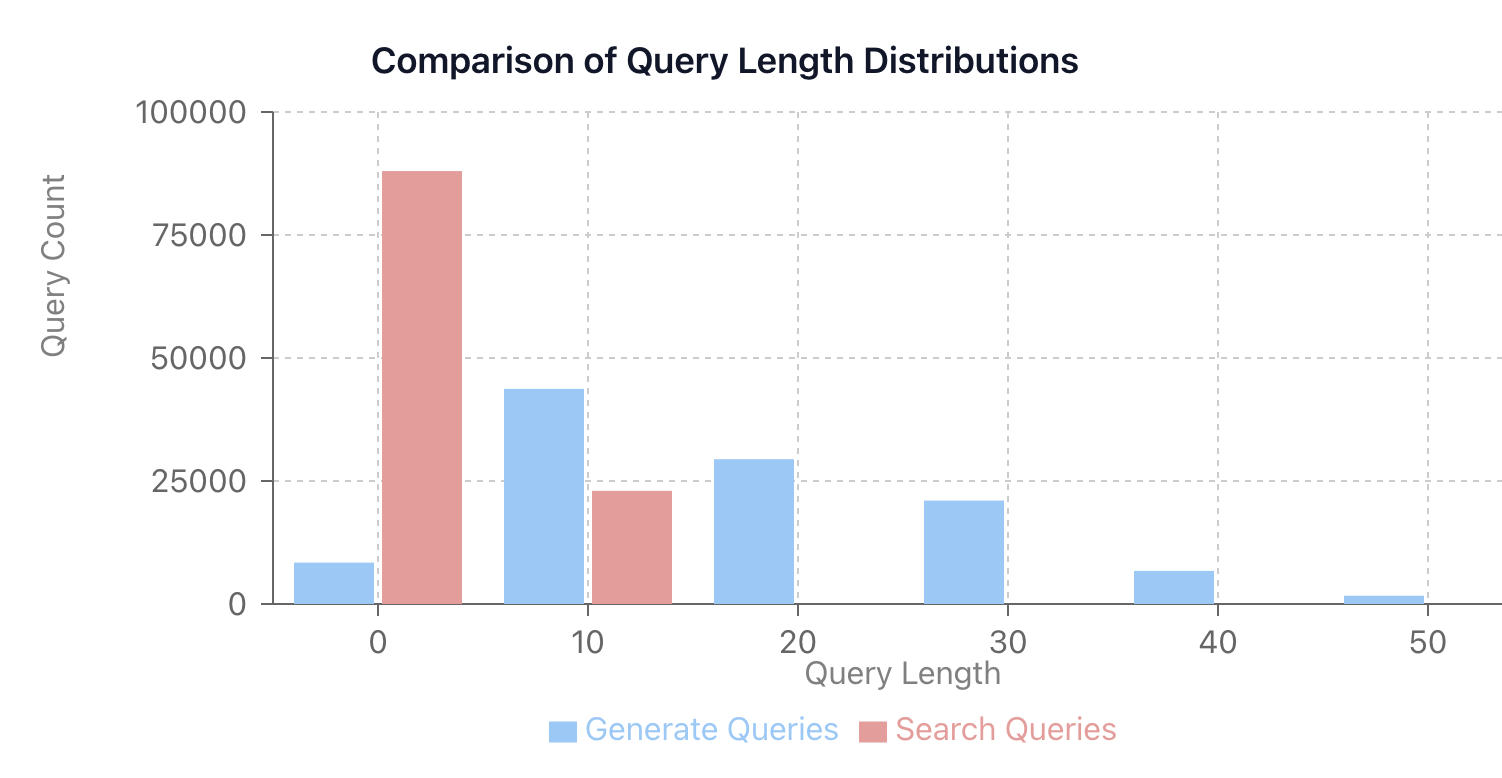}
    \caption{Comparison of search and generate query length}
    \label{fig:search_query_length}
\end{figure}

To ensure our dataset accurately reflects real-world query variations, we designed distinct length distributions for ``Search'' and ``Generate'' queries.  We observed that users tend to use the ``Search'' API to look for generic content and then select a result to start their design with. In contrast, they tend to use the ``Generate'' API to create  specific content which may not exist in the content library. Consequently, ``Search'' queries are usually shorter than ``Generate'' queries.  By accommodating a spectrum of query lengths, our approach improves the model's ability to handle both concise and complex user requests. For ``Search'' based queries, we upper-bound the length to 10 words. This allows the model to focus on short, targeted requests, enhancing retrieval performance. In contrast, ``Generate'' based queries allow  a broader range of lengths, with an upper limit set to 40 words (Figure~\ref{fig:search_query_length}). By expanding the length allowance for ``Generate'' queries, the model learns to handle more descriptive inputs, improving its ability to create content that aligns with nuanced user specifications.

\subsection{Performance Metrics and Evaluation}

\begin{table*}[t]
\centering\small
\begin{tabular}{lccc}
\toprule
\textbf{Model} & \textbf{Function Call} & \textbf{ContentType} & \textbf{Subprompt} \\
\textbf{} & \textbf{F1-Score} & \textbf{Accuracy (CTA)} & \textbf{Similarity (SS)} \\ \midrule
Vanilla Gemma2-2B-Instruction-Tuned &  0.626 & 0.337 & 0.882 \\ 
Fine-tuned Gemma2-2B-Instruction-Tuned & 0.876 & 0.552 & 0.91 \\
\midrule
Vanilla Qwen2.5-1.5B-Instruct & 0.687 & 0.274 & 0.796 \\ 
Fine-tuned Qwen2.5-1.5B-Instruct & 0.863 & 0.554 & 0.91 \\ 
\midrule
Vanilla Qwen2.5-0.5B-Instruct & 0.187 & 0 & 0.02 \\ 
Fine-tuned Qwen2.5-0.5B-Instruct & 0.876 & 0.554 & 0.91 \\ 
\midrule
Vanilla Phi-3.5-mini-Instruct & 0.626 & 0.406 & 0.915 \\ 
Fine-tuned Phi-3.5-mini-Instruct & 0.889 & 0.576 & 0.91 \\ 
\midrule
Vanilla Llama-3.2-1B-Instruct & 0.432 & 0 & 0.182 \\ 
Fine-tuned Llama-3.2-1B-Instruct  & 0.865 & 0.57 & 0.91 \\ 
\bottomrule
\end{tabular}
\caption{Performance summary of the additional SLM models before/after training on the router-based synthetic dataset. The ContentType Accuracy and Sub-prompt Similarity are referenced as CTA and SS respectively}
\label{tab:model_performance_slm}
\end{table*}

\subsubsection{Golden Dataset Details}
To  evaluate model performance, we created a manually curated golden dataset consisting of 460 real-world user queries with high-quality labels. This dataset provides a balanced representation across  query types and intents, with 237 Search queries and 223 Generate queries. The dataset exhibits natural language variation with query lengths ranging from brief phrases to detailed specifications (median length = 8 words, mean = 10.9 words, maximum = 38 words).
This diverse distribution ensures comprehensive evaluation across all supported content types, with particular emphasis on commonly requested media like templates and images while maintaining representation of specialized content types.

\begin{table}[!h]
\centering\small
\begin{tabular}{lrr}
\toprule
\textbf{Content Type} & \textbf{Count} & \textbf{Percentage} \\
\midrule
Templates & 103 & 22.4\% \\
Photos/Images & 97 & 21.1\% \\
Audio & 20 & 4.3\% \\
Video & 19 & 4.1\% \\
Background & 20 & 4.3\% \\
Design Assets & 17 & 3.7\% \\
Text & 20 & 4.3\% \\
Any (type-agnostic) & 164 & 35.7\% \\
\bottomrule
\end{tabular}
\caption{Content Type Distribution in Golden Dataset}
\label{tab:content-distribution}
\end{table}

Each query in the dataset is annotated with function classification (Search/Generate), content type label and optimized sub-prompt.The dataset includes queries ranging from simple search requests to complex generation specifications, as illustrated in Table~\ref{tab:example-queries}. This variation in query complexity and specificity allows us to assess model performance across different difficulty levels and use cases. The content type distribution in golden dataset is  in Table~\ref{tab:content-distribution}. This comprehensive labeling enables evaluation across multiple dimensions of model performance, from high-level task classification to the nuanced understanding required for subprompt optimization. The golden dataset served as our primary benchmark for comparing different model variants and synthetic data generation approaches, providing consistent and reliable metrics for Function Call F1 score, Content Type Accuracy (CTA), and Subprompt Similarity (SS) as reported in Tables~\ref{tab:model_performance} and~\ref{tab:model_performance_slm}.

\subsubsection{Model Performance Metrics}
We began with the baseline Gorilla openfunctions v2 model \cite{c22} fine-tuned for API calls, which yielded initial F1-Score of 0.646, Content Type Accuracy (CTA) of 0.239, and Subprompt Similarity (SS) of 0.824 (Table \ref{tab:model_performance}). These metrics highlighted areas for improvement to meet our query mapping requirements. Testing a single-prompt response approach resulted in F1-Score of 0.788, CTA of 0.57, and SS of 0.898, indicating modest gains but underscoring the need for additional fine-tuning strategies.

To enhance performance further, we fine-tuned the model on a heuristic dataset from our storage index, which included captions, template phrases (e.g.\ "birthday template" for birthday-related queries), and keywords based on query analysis. As a result, F1 rose to 0.801, CTA to 0.676, and SS to 0.919. Next, we implemented the multi-prompt router approach, generating synthetic data using domain-specific metadata from our KG. Fine-tuning on this synthetic dataset alone achieved F1 of 0.844, CTA of 0.65, and SS of 0.867, showing the effectiveness of our router-based approach in capturing nuanced query patterns and context. Combining the synthetic and heuristic datasets yielded further improvements, with F1 reaching 0.875, CTA at 0.737, and SS at 0.915, demonstrating the benefits of blending structured metadata with generated queries. Finally, we applied prompt-tuned fine-tuning to the combined dataset, achieving peak results: F1 of 0.881, CTA of 0.756, and SS of 0.918. This iterative process validated the model's capacity to handle complex queries. To assess whether the observed gains in Function Calling Accuracy between the Single Prompt Fine-Tuned Gorilla model and the Prompt Tuned Gorilla model (Synthetic + Heuristic) dataset using the router were statistically significant, we conducted McNemar's test. The p-value  of 2.529 e-05 demonstrates a highly significant difference. For Subprompt Similarity (SS) metric, a paired t-test yielded a p-value of 0.064, suggesting a trend toward significance.

In addition to training on Gorilla, we  compared the performance of Small Language Models (SLMs) before and after being trained on our router-based synthetic dataset. The SLMs used for comparison were Gemma2-2B-it (Instruction-Tuned) model~\cite{c30}, Qwen2.5-1.5B-instruct and Qwen2.5-0.5B-instruct models~\cite{c31}, Phi3.5-mini-instruct model~\cite{c32} and Llama-3.2-1B-Instruct model~\cite{c33}. We found a significant improvement in the Function Call F1 score, CTA and SS scores across all the SLMs after fine-tuning them with the Router-based synthetic dataset. Additional details on performance improvement are provided in Table~\ref{tab:model_performance_slm}.

\section{Conclusion}
LLMs are fueling efforts to develop systems that accurately interpret user queries and map them to  function calls. However, the scarcity of real-world user data and privacy constraints on training with it necessitate synthetic data generation. Existing synthetic data generation approaches  lack the diversity and complexity needed to mirror real-world interactions, limiting model performance. We introduced a novel architecture for generating high-quality synthetic training data. Our approach integrates content metadata and domain-specific KGs with text-to-text and vision-to-text models,  producing more varied and representative data. Through iterative development, we arrived at a router-based multi-modal architecture that enhances data diversity and improves model training outcomes. Our model demonstrates  gains in function mapping accuracy, although further improvement is possible in content-type classification. 

\section{Future Work}
This research opens several promising avenues for future investigation. One primary direction is to extend the system's linguistic capabilities to support multilingual query processing, thereby improving global accessibility. Although our architecture has proven effective in digital content creation, its underlying principles could be generalized to other domains requiring sophisticated function-calling mechanisms. Leveraging more advanced language models—such as Llama-405B \cite{c33} or DeepSeek \cite{c36}—for synthetic data generation may yield higher-quality training examples, while expanding our golden dataset could enable more rigorous model evaluation. Additionally, exploring the architecture's extensibility to support additional specialized functions and API calls would both broaden the system's applications and provide insights into the scalability of our approach across different functional domains.


\bibliography{acl}

\newpage
\appendix
\section{Model Prompts}
\label{app:appendix_model_prompts}

\subsection{Llama-3.1-70B-Instruct Model Prompts}
\label{app:appendix_llama_model_prompts}

Below is an example of System Prompt used for generating Search data using Llama's 70B model:
\begin{promptbox}
\textbf{Role:} System

\textbf{Content:} You are an AI Assistant responsible for generating a single, concise user search query based on provided metadata. The search queries are short and crisp and less than 10 words. You will be working with different assets for example (templates, images, videos, design assets, backgrounds, shapes). Help me write a search query for an Instagram story template for title:\{title\} focusing on intents:\{intents\}. The query should directly reflect the relevant title, intents, actions, or assets, without any additional explanations or unnecessary text. Do not include any introductory phrases or conclusions, just the query itself.

...

\textbf{Role:} System

\textbf{Content:} You are an AI Assistant responsible for generating a single, concise user search query based on provided metadata. The search queries are short and crisp and less than 10 words. You will be working with different assets for example (templates, images, videos, design assets, backgrounds, shapes). Help me write a search query for the vibrant background for title:\{title\} focusing on actions:\{actions\}. Please include the word background in the query.The query should directly reflect the relevant title, intents, actions, or assets, without any additional explanations or unnecessary text. Do not include any introductory phrases or conclusions, just the query itself.

....

\end{promptbox}

Here are examples of some of the prompts used by Llama 70B model to synthesize Generate function data:
\begin{promptbox}
\textbf{Role:} System

\textbf{Content:} You are an AI that generates creative and engaging user prompts based on provided metadata. The prompt should be less than 40 words. Design a Facebook post prompt for title:\{title\} that encourages users to actions:\{actions\}. Use assets:\{assets\} to support intents:\{intents\}.The prompt should feel like something a human would write and should not include any hashtags or links or unnecessary punctuations.

....

\textbf{Role:} System

\textbf{Content:} You are an AI that generates creative and engaging user prompts based on provided metadata. The prompt should be less than 40 words. Make some prompt for title:\{title\} with intents:\{intents\}. Use assets:\{assets\}, or maybe not?

....

\end{promptbox}

\subsection{InternVL 40B Model Prompt}
\label{app:appendix_internvl_model_prompts}

\begin{promptbox}
\textbf{Role:} System

\textbf{Content:} Based on this image, generate 2 single-sentence prompts that could have created this template. Each prompt should specify the type of material, the purpose it is for, and briefly mention key elements to include. Mention specific business name only if it is present in the image. Translate any non English sentences/words to English.
\end{promptbox}

\section{Model fine-tuning details}
\label{app:appendix_model_finetuning}
For fine-tuning all models, we employed Quantized Low-Rank Adaptation (QLoRA) with consistent hyperparameters across our experiments. The training process utilized a learning rate of $1e^{-4}$ with the AdamW optimizer and cosine learning rate scheduler. We implemented gradient accumulation with 2 steps and a batch size of 4, processing sequences up to 4,096 tokens in length. The models were trained for 3 epochs with a warmup ratio of 0.03, and we applied gradient clipping with a maximum norm of 0.3 to ensure training stability.The LoRA configuration maintained consistency across all models, employing a rank of 16 with an alpha value of 32 and a dropout rate of 0.05. The adaptation targeted key transformation matrices including query, key, value, output, gate, up, and down projections. To optimize memory usage while preserving model quality, we implemented 4-bit quantization (NF4) with double quantization enabled. The training pipeline incorporated mixed precision (FP16) computation and gradient checkpointing for efficient resource utilization. Model evaluation and checkpoint saving were performed at regular intervals of 1,000 steps, with training metrics logged every 20 steps. For inference, we deployed models using vllm with carefully tuned sampling parameters. The configuration included a maximum token length of 4,096, a temperature of 0.3 for controlled randomness, and standard top-k and top-p values of 50 and 1.0 respectively. Each prompt generated a single sample to maintain consistency in our evaluation process. All experimental metrics, including training loss, validation metrics, model checkpoints, and system resource utilization, were tracked and logged using Weights \& Biases (Wandb) for comprehensive experiment monitoring and reproducibility. To ensure reproducibility across all experiments, we maintained a fixed random seed of 42 throughout both training and inference phases. 

\begin{figure}[!htbp]
    \centering
    \includegraphics[width=0.4\textwidth]{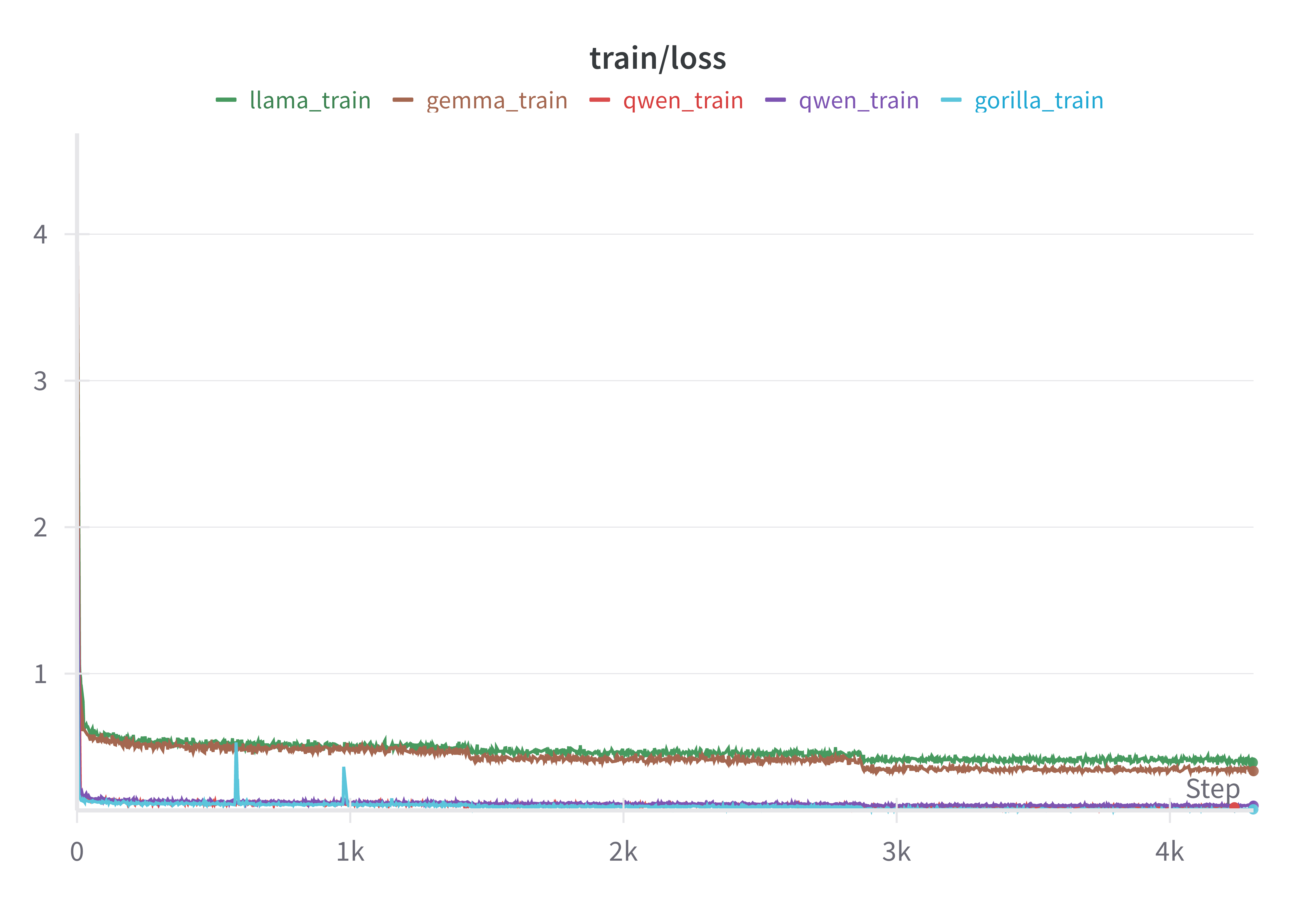}
    \caption{Training loss comparison for: Gorilla, Gemma, Qwen (both variants) and Llama models}
    \label{fig:train_loss_comps}
\end{figure}

\begin{figure}[!htbp]
    \centering
    \includegraphics[width=0.4\textwidth]{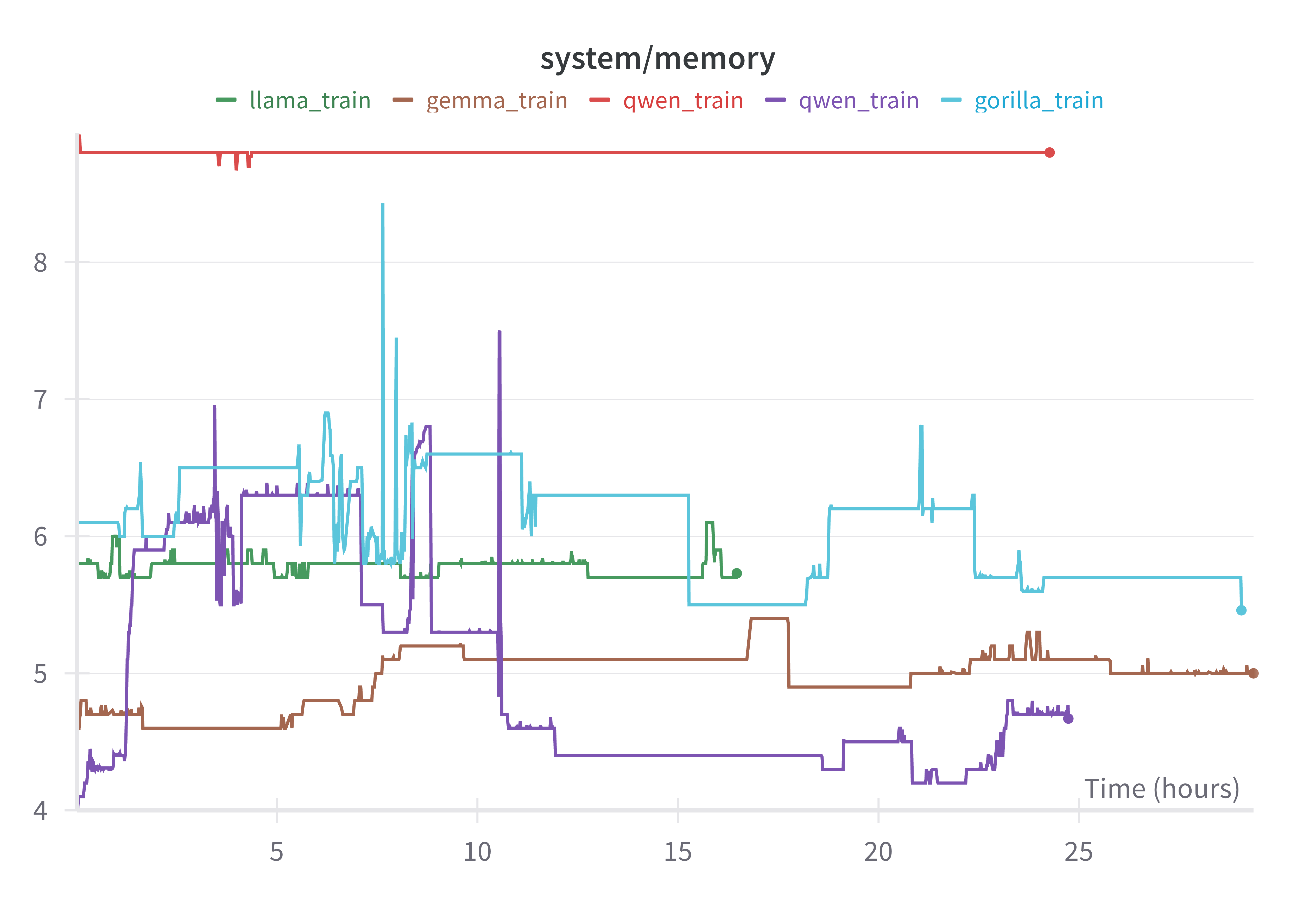}
    \caption{System memory utilization comparison for: Gorilla, Gemma, Qwen (both variants) and Llama models}
    \label{fig:sys_mem_comps}
\end{figure}

\begin{figure}[!htbp]
    \centering
    \includegraphics[width=0.4\textwidth]{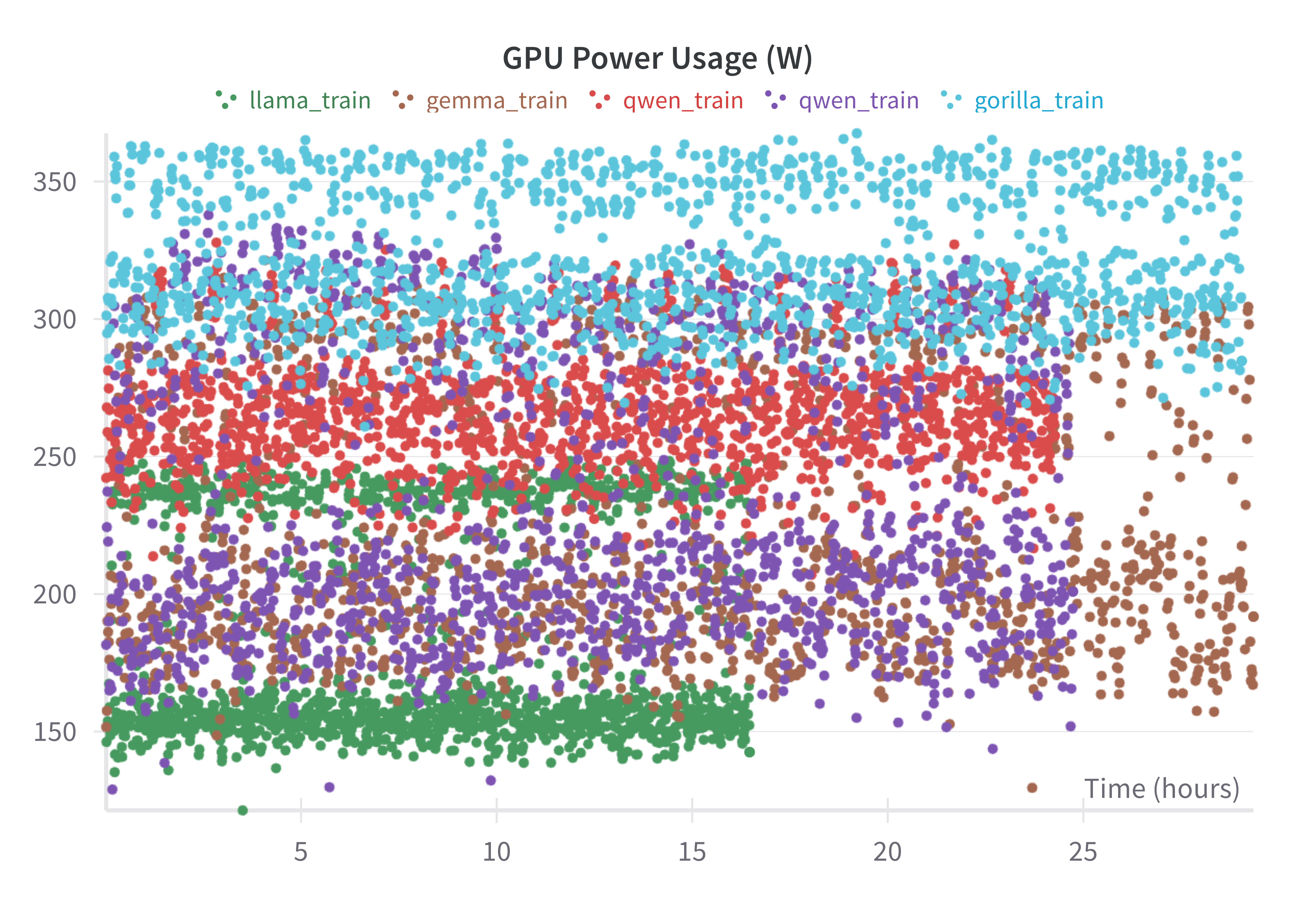}
    \caption{Process GPU utilization comparison for: Gorilla, Gemma, Qwen (both variants) and Llama models}
    \label{fig:gpu_use_comps}
\end{figure}

The following plots capture the comparison of the training loss (Figure ~\ref{fig:train_loss_comps}), system memory utilization (Figure~\ref{fig:sys_mem_comps}) and GPU utilization (Figure~\ref{fig:gpu_use_comps}) for the following models: Gorilla, Gemma2-2B-it model, Qwen2.5-1.5B-Instruct, Qwen2.5-0.5B-Instruct model and Llama3.2-1B-Instruct models.

\subsection{Finetuning Data Structuring and Prompt Preparation}
Before feeding data to the model, we pre-structure the input using Hugging Face's \texttt{apply\_chat\_template} function. This function organizes the conversation into a list of messages with defined roles (e.g.\ \texttt{system} and \texttt{user}), ensuring that the prompt adheres to the format expected by the model. In our implementation, the prompt is prepared in two parts: one describing the task and another providing the actual query and function descriptions.

The prompt template is defined as follows:

\begin{promptbox}
\textbf{Role:} System

\textbf{Content:}  You are an expert in composing functions. You are given a set of possible functions and a question. Based on the question, you will need to make one function/tool call to achieve the purpose. You should only return the function call in your response. You MUST put it in the format of func\_name(params\_name1=params\_value1, params\_name2=params\_value2...). You SHOULD NOT include any other text in the response.

\textbf{Role:} User

\textbf{Content:}<<function>>{function\_descriptions}<<question>>{query}
\end{promptbox}

The following Python code snippet demonstrates how the prompt is generated and tokenized before being passed to the model:

The \texttt{apply\_chat\_template} function performs several key tasks:
\begin{enumerate}
    \item \textbf{Input Organization:} It takes a list of messages, each tagged with a role (either \texttt{system} or \texttt{user}), and concatenates them into a single input string that respects the intended conversational format.
    \item \textbf{Tokenization:} The function tokenizes the structured messages, converting them into a format suitable for the model.
    \item \textbf{Generation Prompt Addition:} It appends any necessary generation prompts that guide the model's response.
    \item \textbf{Tensor Conversion:} Finally, the tokenized data is converted into tensors (using \texttt{return\_tensors="pt"} for PyTorch), ensuring compatibility with the model's expected input format.
\end{enumerate}

This preprocessing step is critical for maintaining the structure and consistency of the input data, thereby facilitating effective fine-tuning and ensuring that the model generates outputs that align with the desired format.

\end{document}